\documentclass[sigconf]{acmart}
\AtBeginDocument{%
  }

\usepackage{amsmath}
\usepackage{subcaption}
\usepackage{multirow}
\usepackage{multicol}
\usepackage{colortbl}
\definecolor{mygray}{gray}{.9}
\usepackage{color}
\usepackage{xcolor} 
\usepackage{booktabs} 
\usepackage{pifont}
\newcommand{\cmark}{\ding{51}}%
\newcommand{\xmark}{\ding{55}}%

\definecolor{tabhighlight}{rgb}{0.88,0.95,1}
\newcommand{\chl}{\cellcolor{tabhighlight}}

\copyrightyear{2026}
\acmYear{2026}
\setcopyright{cc}
\setcctype{by}
\acmConference[MM '26]{Proceedings of the 34th ACM International Conference on Multimedia}{November 10--14, 2026}{Rio de Janeiro, Brazil}
\acmBooktitle{Proceedings of the 34th ACM International Conference on Multimedia (MM '26), November 10--14, 2026, Rio de Janeiro, Brazil}
\acmDOI{10.1145/3767308.3836534}
\acmISBN{979-8-4007-2213-4/2026/11}

\begin{document}

\title{Prototype Matters: Modality-unified Prototype Self-distillation for Unsupervised Visible-infrared Person Re-identification}

\author{Menglin Wang}
\affiliation{%
  \institution{Nanjing Normal University}
   \city{Nanjing}
  \country{China}}
\email{lynnwang6875@gmail.com}

\author{Xiaojin Gong}
\affiliation{%
  \institution{Zhejiang University}
   \city{Hangzhou}
  \country{China}}
\email{gongxj@zju.edu.cn}

\renewcommand{\shortauthors}{Menglin Wang and Xiaojin Gong}

\begin{abstract}
Estimating reliable cross-modality association is crucial to unsupervised visible-infrared person re-ID. While optimal transport is shown to be a practical solution for cross-modality association, it suffers from the rigidness of hard label assignment without considering the impact of cluster noise. Moreover, enforcing only cross-modality contrast is also suboptimal, as it fails to jointly optimize the similarity relation within and across modality. In this paper, we propose a novel framework for cross-modality learning by well exploitation of prototypes: First, instead of contrasting with cross-modality prototypes, we show that modality-unified prototypical contrast facilitates better modality invariance by jointly and simultaneously optimizing similarity relation within and across-modality. Taking self-prototype as a steady teacher, we further refine the instance-prototype online relation through prototype-guided self-distillation. The two components are optimized in a unified framework, leading to a simple yet effective model. On standard VI-ReID benchmarks, we perform extensive comparison and analysis, validating the effectiveness of our proposed method. Code is available at: https://github.com/Terminator8758/PoSeD.
\end{abstract}

\ccsdesc[500]{Information systems~Top-k retrieval in databases}

\keywords{Person re-identification, Prototype, Distillation learning, Unsupervised learning}


\maketitle

\section{Introduction}
\label{sec:intro}

One of the research focus of person re-identification (re-ID) in recent years is to enable day-to-night person retrieval. To achieve this, visible-infrared re-ID is proposed, matching person images captured by visible cameras in the daytime, with images captured by infrared cameras at night or low-light conditions. Existing VI re-ID methods have achieved promising performance by designing modality-specific network architectures~\cite{sysu17, ye2019modality, ye2020dynamic}, introducing color-based data augmentations to mitigate modality gap~\cite{CAJ, caj_extend, RLE}, or leveraging textual modality~\cite{yu2025csdn, hu2024tvi} to provide assistance in visual representation. Despite the progress, the supervised methods rely on costly annotation within and across modalities, limiting their practical application. As a result, unsupervised visible-infrared re-ID (UnVI-reID) has received increasing attention recently.

Compared to supervised scenario, the main challenge in UnVI-reID is the estimation of reliable intra- and cross-modality associations under significant modality discrepancy~\cite{sysu17, D2RL}. Following the practice of single-modality unsupervised re-ID~\cite{unsup_clustering, clustercontrast, ge2020self}, paradigms like iterative clustering and fine-tuning can be readily applied for intra-modality learning in UnVI-reID. The cross-modality learning, however, struggles in obtaining reliable association due to the modality gap. A commonly adopted strategy is to match the intra-modality obtained clusters through optimal transport based algorithms~\cite{PGMAL} such as Hungarian matching, then utilize the matching result for cross-modality optimization. However, such learning paradigm presents two critical limitations:

First, the intra-modality obtained clusters are inevitably noisy, rendering the cross-modality hard matching assignment suboptimal. Intuitively, considering the noise within clusters, a visible cluster should be matched to infrared clusters with \textit{soft} probability according to their semantic consistency. Moreover, as the cluster prototypes are updated on-the-fly during batch-wise training, their semantic relations might change and drift from the early offline matching, necessitating a more reliable association strategy that perceives online prototype similarity dynamics.

Second, the cross-modality feature optimization is another issue worth attention. Due to modality gap, the cross-modality similarity is significantly lower than intra-modality similarity. Commonly-utilized cross-modality contrastive loss only pulls the instance closer to its matched cross-modality prototype compared to other cross-modality prototypes. Without explicitly addressing intra- and cross-modality similarity discrepancy, the generalization ability of learned representation is likely to be compromised when facing severe modality shift.

\begin{figure}[ht]
\centering
\includegraphics[width=0.5\textwidth]{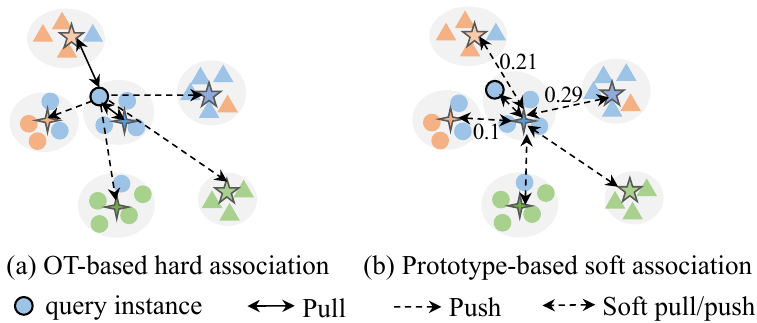}
\caption{Illustration on the motivation of modality-unified prototype self-distillation. Different color represents different identity, and different shape denotes different modality.}
\label{fig_intro}
\end{figure}

Given the cross-modality similarity discrepancy, we argue that modality-unified contrastive optimization that explicitly considers the modality gap is a better alternative for cross-modality optimization. Instead of only optimizing the similarity between an instance and its cross-modality prototypes, we propose to jointly optimize the instance's similarity with both intra- and cross-modality prototypes. The main advantage of modality-unified optimization is allowing the model to simultaneously perceive the similarity disparity within and across-modality, so that the model learns to optimize cross-modality semantic relations with reference to intra-modality similarity distribution. Through such optimization, the model is forced to perceive and reduce the cross-modality similarity gap, thus achieving modality invariance while seeking for identity discrimination.

To rectify the instance-to-prototype association, we draw inspiration from the self-distillation mechanism in self- and semi-supervised learning~\cite{grill2020bootstrap, caron2021emerging, sohn2020fixmatch, wen2023parametric}, and propose a \textit{prototype-driven self-distillation} strategy. Self-distillation has proven to be an effective mechanism for providing supervision from the unlabeled data itself. A common self-distillation strategy is to regard two randomly augmented views of the same image as each other's teacher, then use one view's prediction as the distillation target for the other. However, the \textit{intra-instance variance} is limited to pre-defined augmentations, and does not enhance the learning of \textit{intra-identity semantics} which is crucial to identity discrimination. 

To better adapt self-distillation into UnVI-reID, we take a look at the online-updated prototypes, and find the centroid prototypes to be a steady-evolving yet more informative substitute for distillation. As centroid prototypes are online updated by intra-cluster instances, they harness the instances variation and reflect the intra-cluster distribution. Therefore, we regard each instance's self-prototype as the teacher, and derive an instance-adaptive similarity distribution as the online soft distillation target for the instance in consideration. Compared to the offline optimal-transport (OT) based association, the online distillation rectifies instance-to-prototype association in a soft way, as shown in Figure \ref{fig_intro}. Moreover, we build the prototype-guided self-distillation upon the proposed modality-unified optimization paradigm, so that the distillation can be integrated for different modalities through a modality-aware normalization.

The online association driven by the prototype self-distillation complements the inefficiency of offline prototype matching, leading to robust cross-modality association estimation. We unify the offline matching and online distillation within the proposed unified contrastive framework, thereby facilitating modality-invariant yet ID-discriminative representation learning. 

To summarize, our main contribution are as follow:
\begin{itemize}
  \item We propose a \textit{simple yet effective} framework for UnVI-ReID, offering a new perspective to exploit prototypes for improving association and model representation learning in the face of modality discrepancy. 
  \item We propose modality-unified contrastive loss to explicitly address the cross-modality similarity discrepancy. Building on the modality-unified optimization paradigm, we further design a prototype-driven self-distillation loss to rectify the noise of offline cross-modality matching, by taking prototype as a steady and strong teacher. 
  \item Extensive comparison and ablations on standard VI-reID benchmarks validate the effectiveness of the proposed method.
\end{itemize}

\section{Related Work}

\subsection{Unsupervised Visible-infrared Re-ID}
Considering the distribution discrepancy of visible and infrared modalities~\cite{sysu17,D2RL}, current UnVI-reID methods usually incorporate both intra-modality  and cross-modality learning. For intra-modality learning, most methods follow the iterative clustering framework which has proven effective in single-modality unsupervised re-ID~\cite{unsup_clustering, clustercontrast, ge2020self, Wang2021CAP}. Building on this framework, RPNR~\cite{RPNR} improves pseudo label by selecting reliable instances within cluster for label re-assignment. NULC~\cite{teng2024ULC} leverages the correlation between nearest neighbors and clusters to calibrate the cluster labels and generate loss weight for each instance. CCLNet~\cite{CCLNet} injects text semantics by learning pseudo text prompts to improve visual representation learning. PCLHD~\cite{PCLMP} enhances representation learning by designing complementary types of prototypes to capture clusters divergence and variety.

Cross-modality learning focus on improving either cross-modality association quality or representation learning effectiveness. For the former, PGM~\cite{PGMAL} proposes a multi-step optimal transport algorithm to estimate cross-modality correspondence. GUR~\cite{yang2023GUR} establishes cross-modality relation by intra-modality label smoothing and cross-modality label propagation. ADCA~\cite{ADCA} associates cross-modality clusters by counting and ranking the matched instance pairs. CAM~\cite{CAM2025} improves cross-modality cluster similarity measure by leveraging channel-augmented visible images for dual-similarity fusion. To improve representation learning, PGM~\cite{PGMAL} designs modality alternation to reduce the impact of noise, SDCL~\cite{SDCL} enforces consistency between shallow and deep feature prediction to improve cross-modal alignment. In this work, we also focus on improving cross-modality learning by addressing both association and representation from the perspective of prototypes.

\subsection{Self-distillation Learning}
In self-supervised~\cite{chen2020simple, grill2020bootstrap, caron2021emerging} and semi-supervised learning~\cite{tarvainen2017mean, berthelot2019mixmatch, sohn2020fixmatch, wen2023parametric}, self-distillation is designed as a way to provide supervision from the unlabeled instance itself. The most common form of self-distillation is generating two random views of the same instance, and using one view's sharpened prediction as the teacher for the other. To introduce more variance between teacher and student prediction, the teacher prediction can be generated from a different network than the student, such as using the EMA model~\cite{tarvainen2017mean, vaze2023no}. Our method also embraces the idea of self-distillation for introducing additional refined supervision for the unlabeled data. Different from other works employing self-augmented views, we uniquely exploit cluster prototype as a steady teacher to provide strong guidance and rectify online instance-to-prototype relations.

\subsection{Learning with Prototypes}
The concept of prototype has been widely applied in many unsupervised learning tasks including person re-ID~\cite{PCLMP, Wang2021CAP, clustercontrast, DCMIP, liubmw}. By designing prototypes that capture fine-grained cluster distribution characteristics, such as camera view variation~\cite{Wang2021CAP} and diversity~\cite{PCLMP}, or improving prototype updating mechanism~\cite{liubmw, CCLNet}, prototypes can be a strong guidance to discriminative representation learning through the optimization of prototypical contrast. Nevertheless, most methods focus on the prototype design, without looking into how prototypes can facilitate pseudo label rectification or semantic refinement. In this work, we take the initiative to exploit prototypes as reliable online guidance for improving instance-to-prototype association, from the perspective of self-distillation.

\section{Methodology}

\subsection{Overview}
\begin{figure*}[ht]
\centering
\includegraphics[width=0.98\textwidth]{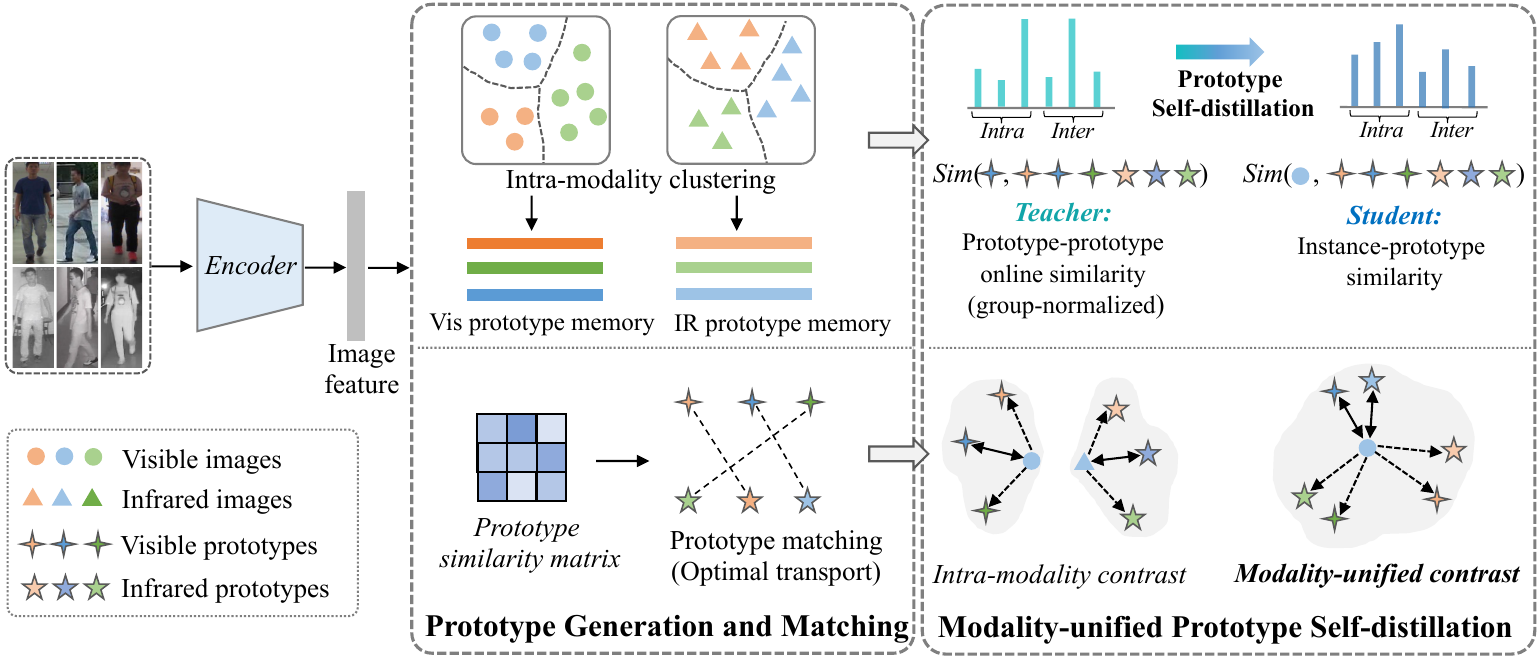}
\caption{An overview of the proposed framework. After intra-modality clustering, visible and infrared prototypes are constructed. Then an optimal transport based algorithm estimates the association of visible and infrared prototypes based on cross-modality prototype similarity. With the matching result, a modality-unified contrast loss is designed to jointly optimize the similarity relation within and across-modality. Then prototype-based online self-distillation refines the matching by introducing prototype-based group-normalized similarity as the soft distillation target.}
\label{fig_framework}
\end{figure*}

Our method focuses on estimating reliable cross-modality association and enhancing modality-invariant representation for UnVI-reID. The overall framework is illustrated in Fig. \ref{fig_framework}. The first training stage involves intra-modality learning where visible and infrared images are separately clustered. Based on cluster pseudo label, visible and infrared prototype memories are initialized and online updated, with the model supervised by intra-modality prototypical contrast. The second stage performs cross-modality learning in a modality-unified manner: optimal transport based algorithm estimates cross-modality matching as the cross-modality hard pseudo label. Further, prototype based self-distillation refines the cross-modality pseudo label by introducing prototype-based online similarity as the soft distillation target. Both hard pseudo label and online distillation are optimized with modality-unified contrastive loss, explicitly aligning cross-modality representation and improving discriminative learning.

\subsection{The UnVI-reID Baseline}
\label{sec:baseline}
In the task of UnVI-reID, suppose the dataset is given as $\mathcal{D}=\mathcal{D}_{v} \cup \mathcal{D}_{ir}$, where $\mathcal{D}_{v}=\{x^v_i\}_{i=1}^{N_{v}}$ is the unlabeled dataset of visible images, and $\mathcal{D}_{ir}=\{x^{ir}_i\}_{i=1}^{N_{ir}}$ is the unlabeled dataset of infrared images. Denote the backbone model as $f()$, in this method we follow previous methods~\cite{ADCA,PGMAL} and use AGW~\cite{AGW} as the backbone network $f()$. Considering the modality discrepancy, separate initial Conv blocks are constructed for visible and infrared images, while the rest network is shared among two modalities. The baseline~\cite{PGMAL} is a two-stage learning method, with the first warmup stage focusing on intra-modality learning, and the second stage incorporating both intra-modality and cross-modality learning. 

\subsubsection{Intra-modality learning.} Following the popular methods~\cite{unsup_clustering, clustercontrast, ge2020self, Wang2021CAP} in single-modality Un-reID, an iterative clustering paradigm is adopted for intra-modality learning. At the beginning of each epoch, unsupervised clustering is performed within visible and infrared modalities separately. Suppose that after clustering, $C_v$ visible clusters and $C_{ir}$ infrared clusters are generated, and pseudo labels for visible image $x^v_i$ and infrared image $x^{ir}_j$ are denoted as $y^v_i$ and $y^{ir}_j$ respectively. 

\noindent \textbf{Dual prototype memory.} To capture the intra-cluster diversity and variance, two types of prototypes~\cite{PCLMP} are constructed for each cluster, \textit{i.e.} a centroid prototype and a hard prototype. The two type of prototypes are both initialized as the average of instance features within the cluster, but differ in the updating mechanism. Take visible modality as the example. During backward propagation, the centroid prototype $M_i^v$ is updated by the average feature of in-batch instances of pseudo label $y^v_i$, while the hard prototype is updated using the the hard positive instance, \textit{i.e.} instance with label $y_i$ and having the largest distance with the prototype $P^v[i]$:
\begin{equation}
\mathcal{M}^v[i] \leftarrow \mu \cdot \mathcal{M}^v[i] + (1-\mu) \cdot mean(f(x^v_k)|y_k=i),
\end{equation}

\begin{equation}
\mathcal{P}^v[i] \leftarrow \mu \cdot \mathcal{P}^v[i] + (1-\mu) \cdot hard(f(x^v_k)|y_k=i),
\end{equation}
where $\mu$ is the updating momentum. $M^v \in R^{C_v\times d}$ represents the centroid prototype memory, and $P^v \in R^{C_v\times d}$ denotes the hard prototype memory.

The image $x^v_i$ is then contrasted with both centroid and hard prototypes separately, leading to the intra-modality prototypical contrastive loss as below:
\begin{equation}
\begin{split}
\mathcal{L}^v_{intra} &= -\sum_{i=1}^{B} \log \frac{exp(\mathcal{M}^v[y_i]^T f(x^v_i)/\tau)}{\sum_{j=1}^{C_v} exp(\mathcal{M}^v[j]^T f(x^v_i)/\tau)} \\
&\quad  -\sum_{i=1}^{B} \log \frac{exp(\mathcal{P}^v[y_i]^T f(x^v_i)/\tau)}{\sum_{j=1}^{C_v} exp(\mathcal{P}^v[j]^T f(x^v_i)/\tau)},
\label{eq_intra_loss}
\end{split}
\end{equation}
where $B$ is the batch size, and $\tau$ is the temperature. By optimizing the intra-modality contrastive loss, the image is pulled closer to its belonging centroid and hard prototypes, while pushed away from the other negative prototypes, thus achieving intra-modality discrimination. 

For the infrared modality, the prototypes ($\mathcal{M}^{ir}$, $\mathcal{P}^{ir}$) and intra-modality loss $\mathcal{L}^{ir}_{intra}$ can be formulated similarly. Then the total intra-modality loss is computed as: $\mathcal{L}_{intra}=\mathcal{L}^v_{intra}+\mathcal{L}^{ir}_{intra}$.

\subsubsection{Cross-modality learning.} With the intra-modality clustering result and the obtained prototypes, the next step is associating the intra-modality clusters so that cross-modality correspondence can be established. Given the initial cluster mean features $\tilde{\mathcal{M}^v}$ and $\tilde{\mathcal{M}^{ir}}$, the cross-modality cluster similarity matrix $S \in R^{C_v\times C_{ir}}$ can be computed as the cosine similarity between pairwise features:
\begin{equation}
S[i, j]=\frac{\tilde{\mathcal{M}^v}[i]\cdot \tilde{\mathcal{M}^{ir}}[j]}{||\tilde{\mathcal{M}^v}[i]|| \cdot ||\tilde{\mathcal{M}^{ir}}[j]||},
\end{equation}
where $||\cdot||$ denotes the $L_2$ norm. Then the cost matrix can be derived as $Cost[i,j]=1/exp(S[i,j])$.

Following the multi-step matching strategy of PGM~\cite{PGMAL}, the cross-modality correspondence ($V2R, R2V$) can be predicted by minimizing the matching cost. Then based on the correspondence result, the cross-modality contrastive loss is computed between image $x^v_i$ and its cross-modality prototypes:
\begin{equation}
\begin{split}
\mathcal{L}^v_{cross} &= -\sum_{i=1}^{B} \log \frac{exp(\mathcal{M}^{ir}[V2R[y_i]]^T f(x^v_i)/\tau)}{\sum_{j=1}^{C_{ir}} exp(\mathcal{M}^{ir}[j]^T f(x^v_i)/\tau)}\\
& -\sum_{i=1}^{B} \log \frac{exp(\mathcal{P}^{ir}[V2R[y_i]]^T f(x^v_i)/\tau)}{\sum_{j=1}^{C_{ir}} exp(\mathcal{P}^{ir}[j]^T f(x^v_i)/\tau)}.
\label{eq_cross_loss}
\end{split}
\end{equation}

Minimizing $\mathcal{L}^v_{cross}$ encourages the image to be more similar to its cross-modality matched prototypes compared to the rest cross-modality prototypes, so that the model learns to enhance its ability to recognize cross-modality identities.

\subsection{Modality-unified Prototypical Contrast}
\label{sec:unified}
Although the cross-modality loss in Eq. \ref{eq_cross_loss} is an intuitive formulation for cross-modality learning, it only optimizes the \textit{relative similarity} between an image and the cross-modality prototypes, without explicitly addressing the cross-modality similarity gap caused by modality discrepancy. As a result, the overall intra-modality similarities remain larger than cross-modality similarities. Ideally, we would expect the representation to be robust to modality shift, so that the model can focus on truly discriminative cues shared among modalities. To this end, we propose a modality-unified prototypical contrastive loss that jointly perceive and optimize intra- and cross-modality similarity within a single loss.

\noindent \textbf{\textit{Unified pseudo label.}} Given the cross-modality cluster matching result, we first transform it into unified pseudo label for each visible or infrared cluster. When the number of visible and infrared clusters are not equal, a multi-step matching leads to multiple clusters matched to the same cross-modality cluster. We assign the same pseudo label to the matched cross-modality clusters regardless of the order they are matched.

\noindent \textbf{\textit{Unified prototype memory.}} By concatenating the visible and infrared prototypes, we obtain a unified prototype memory. Suppose pseudo label $z_k$ is assigned to the $k$-th unified prototype. Then the modality-unified prototypical loss is computed as a multi-positive contrastive loss as follows:
\begin{equation}
\small
\mathcal{L}^v_{unified} = -\sum_{i=1}^{B} \frac{1}{|pos(y_i)|} \sum_{j \in pos(y_i)} \log \frac{exp(\mathcal{P}[j]^T f(x^v_i)/\tau)}{\sum_{k=1}^{C_v+C_{ir}} exp(\mathcal{P}[k]^T f(x^v_i)/\tau)},
\label{eq_unified_loss}
\end{equation}
where $\mathcal{P}$ is the unified hard prototype memory, and $pos(y_i)$ denotes the prototypes sharing the same unified pseudo label as prototype $y_i$. Note that we only optimize the modality-unified contrast using the hard prototype memory, since we empirically find the centroid prototypes contribute little to unified optimization when employing the OT-based hard association.

\subsection{Prototype-guided Self-distillation}
\label{sec:distill}
As prototypes are online updated with batch instances, their semantics may drift from the initial state and the offline matching becomes suboptimal. Moreover, the clusters are inherently noisy, so the hard label assignment may not best reflect the semantic relation between instance and clusters. Inspired by the self-distillation mechanism in self-supervised learning~\cite{chen2020simple, grill2020bootstrap, caron2021emerging,berthelot2019mixmatch}, \textbf{we consider each instance's belonging prototype as an online teacher for similarity distillation.} Specifically, taking the centroid prototype as a steady-evolving representation of the cluster semantic, we distill the prototype-to-prototype similarity to the instance-to-prototype similarity.

\noindent \textbf{\textit{Instance-adaptive similarity fusion.}} Denote the prototype-to-prototype similarity of image $x^v_i$ as $Q_{y_i}=\mathcal{M}[y_i]^T \mathcal{M}$. Considering the cluster noise, images in the same cluster may not always share consistent semantics with the cluster prototype. To account for this, we propose to fuse the online instance-to-prototype similarity with prototype-to-prototype similarity, leading to an instance-adaptive teacher similarity for self-distillation:
\begin{equation}
Q_{y_i} \leftarrow 0.5 \cdot Q_{y_i} + 0.5 \cdot f(x^v_i)^T \mathcal{M}.
\end{equation}

\noindent \textbf{\textit{Modality-aware teacher distribution.}} Following the modality-unified contrastive optimization in Sec. \ref{sec:unified}, we jointly optimize the online similarity \textit{w.r.t.} prototypes of all modalities. Due to the intra- and cross-modality similarity distribution gap, if normalizing the teacher similarity regardless of modality, the intra-modality similarity would dominate, and the cross-modality similarity will be negatively suppressed. To avoid such situation, we propose a modality-aware group normalization strategy: 

Denote the sharpened similarity as $Q_{y_i}/\gamma$, where $\gamma$ is the scaling temperature. To avoid the modality-induced similarity bias, we separately normalize the similarity \textit{w.r.t.} visible prototypes and infrared prototypes, \textit{i.e.} the first $C_v$ elements of $Q_{y_i}$ and the rest elements, deriving the modality-normalized similarity $\bar{Q}_{y_i}$:
\begin{equation}
\bar{Q}_{y_i}=[Softmax(Q_{y_i}[0:C_v]/\gamma);Softmax(Q_{y_i}[C_v:]/\gamma)]. 
\end{equation}
Then another global normalization is performed to ensure the final output is a probability distribution: 
\begin{equation}
\bar{Q}_{y_i} \leftarrow \bar{Q}_{y_i}/sum(\bar{Q}_{y_i}).
\end{equation}

Utilizing the prototype-based teacher distribution $\bar{Q}$ as the soft distillation target, the self-distillation loss is formulated as:
\begin{equation}
\mathcal{L}^v_{sd} = -\sum_{i=1}^{B} \sum_{j=1}^{C_v+C_{ir}} \bar{Q}_{y_i}[j]\log \frac{exp(\mathcal{M}[j]^T f(x^v_i)/\tau)}{\sum_{k=1}^{C_v+C_{ir}} exp(\mathcal{M}[k]^T f(x^v_i)/\tau)}.
\label{eq_distill_loss}
\end{equation}

\noindent \textbf{\textit{Discussion.}}
The concept of self-distillation has been utilized in many semi-supervised and self-supervised learning approaches. However, unlike common strategies using randomly-augmented views to provide teacher-student self-distillation, we \textit{offer a novel perspective of repurposing prototypes as the source of distillation.} In many unsupervised learning tasks especially re-ID, prototypes are an important assistance, but its effect has not been fully understood or exploited yet. In our method, we take online-updated centroid prototypes as a steady yet informative teacher, distilling the instance-to-prototype similarity distribution to provide reliable guidance for instance-prototype similarity optimization.

\subsection{The Overall Loss} 
The modality-unified contrastive loss optimizes instance-to-prototype similarity based on cluster matching, while the self-distillation loss refines online association through prototype-based self-distillation. To exploit their complementarity, we optimize the two losses within a unified framework, and the overall loss is computed as:
\begin{equation}
\mathcal{L}_{overall} =
\begin{cases}
\mathcal{L}^v_{sd}+ \mathcal{L}^{ir}_{sd} + \mathcal{L}^{v}_{unified} & \text{if } epoch \text{\%2==0} \\
\mathcal{L}^v_{sd}+ \mathcal{L}^{ir}_{sd} + \mathcal{L}^{ir}_{unified} & \text{if } epoch \text{\%2==1}
\end{cases}
\label{eq_overall_loss}
\end{equation}
where $\mathcal{L}_{unified}$ alternates between $\mathcal{L}^v_{unified}$ and $\mathcal{L}^{ir}_{unified}$ in consecutive epochs, to avoid cluster matching noise amplification~\cite{PGMAL}.

\section{Experiment}

\subsection{Datasets and Evaluation Protocols}
Following existing UnVI-reID methods~\cite{PGMAL,RPNR,CAM2025}, we evaluate our method on three benchmark VI-reID datasets: SYSU-MM01~\cite{sysu17}, RegDB~\cite{regdb} and LLCM~\cite{DEEN}. 

\noindent \textbf{SYSU-MM01.} The training set of SYSU-MM01 consists of 22,258 visible images and 11,909 infrared images from 395 identities, captured under 4 visible cameras and 2 infrared cameras. The test set consists of 96 identities. Two cross-modality evaluation settings are typically adopted, All Search and Indoor Search, which differ in the gallery set. The Indoor Search mode includes only indoor images as the gallery, while All Search mode uses both indoor and outdoor visible images as gallery.

\noindent \textbf{RegDB} consists of 412 identities, with 10 visible images and 10 infrared images captured for each identity. Due to the smaller dataset scale, 10 trials are conducted and the average performance is reported. In each trial, 206 identities are randomly selected out of all the 412 identities, and the rest identities constitutes the test set. Following common evaluation protocol, both Visible-to-Thermal and Thermal-to-Visible settings are adopted for evaluation.

\noindent \textbf{LLCM} is a challenging dataset comprising 46,767 images from 1,064 identities, with 16,946 visible and 13,975 infrared images. The images are captured by 9 visible cameras and 8 infrared cameras deployed in low-light environments. The training set contains 30,921 images from 713 identities, and the rest 351 identities forms the test set. Both visible-to-infrared and infrared-to-visible modes are considered in performance evaluation.

For evaluation metrics, the Cumulative Matching Curve (CMC), mean average precision (mAP) are reported.

\subsection{Implementation Details}
We adopt the ResNet50-based AGW~\cite{AGW} as the backbone network. Visible and infrared modalities has independent first convolution blocks, while the rest network is shared for the two modality. Following PGM~\cite{PGMAL}, the model is trained in a two-stage manner, with the first stage focused on intra-modality learning, and the second stage covering both intra- and cross-modality learning. Each of the two training stages consists of 50 epochs, and Adam optimizer with an initial learning rate of 0.00035 is utilized for both stages. The learning rate is divided by 10 every 20 epochs. At the beginning of each epoch, DBSCAN is utilized for unsupervised clustering in visible and infrared modality separately, with the clustering threshold $\textit{eps}$ set to 0.6 on SYSU-MM01 and 0.3 on RegDB, following existing methods. For SYSU-MM01, we adopt the subset sampling strategy~\cite{wang2026modality} when clustering on the visible modality to alleviate modality imbalance. For batch sampling, we adopt the common PK sampler, which randomly samples 8 clusters and 16 images per cluster, forming 128 images sampled per modality and 256 images per batch. Data augmentation includes random cropping, random flipping, random grayscale and random colorjitter. The momentum $\mu$ for prototype updating is set to 0.1, and temperature $\tau$ is set to 0.05. The teacher temperature $\gamma$ is set to 0.05 as well.

\begin{table*}[ht]
\centering
\caption{Comparison with state-of-the-art methods on SYSU-MM01 and RegDB dataset. The best unsupervised performance is marked in \textbf{Bold} and the second best is in \underline{underline}. \textbf{Ours}$^\star$ indicates our method combined with BMIL framework.}
\scalebox{1.0}{
\begin{tabular}{c|c|c||c|c|c|c||c|c|c|c}
\bottomrule
\multicolumn{3}{c||}{\multirow{2}{*}{Settings}} & \multicolumn{4}{c||}{SYSU-MM01} & \multicolumn{4}{c}{RegDB} \\ \cline{4-11}
\multicolumn{3}{c||}{} & \multicolumn{2}{c|}{All Search} & \multicolumn{2}{c||}{Indoor Search} & \multicolumn{2}{c|}{Visible to Thermal} & \multicolumn{2}{c}{Thermal to Visible}\\
\hline
Type & Method & Venue & Rank-1 & mAP     & Rank-1 & mAP      & Rank-1 & mAP     & Rank-1 & mAP \\
\hline
\multirow{14}{*}{Sup-VI-ReID}    
                ~ & DDAG~\cite{DDAG} & ECCV'20 & 54.8 & 53.0 & 61.0 & 68.0 & 69.4 & 63.5 & 68.1 & 61.8 \\ 
                ~ & AGW~\cite{AGW} & TPAMI'21 & 47.5 & 47.7 & 54.2 & 63.0 & 70.1 & 66.4 & 70.5 & 65.9 \\  
                ~ & CAJ~\cite{CAJ} & ICCV'21 & 69.9 & 66.9 & 76.3 & 80.4 & 85.0 & 79.1 & 84.8 & 77.8 \\  
                ~ & MPANet~\cite{MPANet} & CVPR'21 & 70.6 & 68.2 & 76.7 & 81.0 & 83.7 & 80.9 & 82.8 & 80.7 \\ 
                ~ & DART~\cite{DART} & CVPR'22 & 68.7 & 66.3 & 72.5 & 78.2 & 83.6 & 75.7 & 82.0 & 73.8 \\
                ~ & FMCNet~\cite{FMCNet} & CVPR'22 & 66.3 & 62.5 & 68.2 & 74.1 & 89.1 & 84.4 & 88.4 & 83.9 \\ 
                ~ & DEEN~\cite{DEEN} & CVPR'23 & 74.7 & 71.8 & 80.3 & 83.3 & 91.1 & 85.1 & 89.5 & 83.4 \\
                ~ & PartMix~\cite{PartMix} & CVPR'23 & 77.8 & 74.6 & 81.5 & 84.4 & 85.7 & 82.3 & 84.9 & 82.5\\                 
                ~ & MUN~\cite{MUN} & ICCV'23 & 76.2 & 73.8 & 79.4 & 82.1 & 95.2 & 87.2 & 91.9 & 85.0\\     
                ~ & RLE~\cite{RLE} & NeurIPS'24 & 75.4 & 72.4  & 84.7 & 87.0 & 92.8 & 88.6 & 91.0 & 86.6\\   
                 ~ & CSDN~\cite{yu2025csdn} & TMM'25 & 75.2 & 71.8 & 82.0 & 85.0 & 89.0 & 84.7 & 88.2 & 82.8 \\
\midrule
\multirow{3}{*}{SemiSup-VI-ReID}
               ~ & OTLA~\cite{OTLA} & ECCV'22 & 48.2 & 43.9 & 47.4 & 56.8 & 49.9 & 41.8 & 49.6 & 42.8\\
               ~ & DPIS~\cite{DPIS} & ICCV'23 & 58.4 & 55.6 & 63.0 & 70.0 & 62.3 & 53.2 & 61.5 & 52.7\\
               ~ & HECC~\cite{zhang2025weakly} & ICCV'25 & 70.4 & 66.6 & 76.5 & 80.2 & - & - & - & - \\
\midrule
\multirow{12}{*}{Unsup-VI-ReID}
               ~ & OTLA~\cite{OTLA} & ECCV'22 & 29.9 & 27.1 & 29.8 & 38.8 & 32.9 & 29.7 & 32.1 & 28.6\\ 
               ~ & ADCA~\cite{ADCA} & MM'22 & 45.5 & 42.7 & 50.6 & 59.1 & 67.2 & 64.1 & 68.5 & 63.8\\   
               ~ & CCLNet~\cite{CCLNet} & MM'23 & 54.0 & 50.2 & 56.7 & 65.1 & 69.9 & 65.5 & 70.2 & 66.7\\
               ~ & PGM~\cite{PGMAL} & CVPR'23 & 57.3 & 51.8 & 56.2 & 62.7 & 69.5 & 65.4 & 69.9 & 65.2\\           
               ~ & GUR~\cite{yang2023towards} & ICCV'23 & 61.0 & 57.0 & 64.2 & 69.5 & 73.9 & 70.2 & 75.0 & 69.9\\
               ~ & MMM~\cite{MMM} & ECCV'24 & 61.6 & 57.9 & 64.4 & 70.4 & 89.7 & 80.5 & 85.8 & 77.0\\
               ~ & PCLHD~\cite{PCLMP} & NeurIPS'24 & 64.4 & 58.7 & 69.5 & 74.4 & 84.3 & 80.7 & 82.7 & 78.4 \\
               ~ & RPNR~\cite{RPNR} & MM'24  & 65.2 & 60.0 & 68.9 &74.4 & 90.9 & 84.7  & 90.1 & 83.2 \\
               ~ & TokenMatcher~\cite{wang2025tokenmatcher} & AAAI'25 & 65.1 & 62.8 & 69.0 & 74.9 & 93.0 & 86.3 & 91.8 & 85.2 \\
               ~ & ASM~\cite{CAM2025} & ICCV'25  & 65.1 & \underline{63.4} & 71.1 &76.9 & 88.2 & 76.7  & 86.9 & 79.2 \\
               ~ & BMIL~\cite{wang2026modality} & AAAI'26 &  \underline{67.1} & 63.1 & \underline{75.0} & \underline{78.6} & 94.3 & 89.1 & 93.6 & 88.5 \\
               ~ & \chl \textbf{Ours} & \chl This paper  & \chl 66.0 & \chl 62.6 & \chl 72.7 & \chl 76.8 & \chl \underline{95.3} & \chl \underline{89.6} & \chl \textbf{94.9} & \chl \underline{89.2} \\
               ~ & \chl \textbf{Ours}$^\star$ & \chl This paper  & \chl \textbf{67.5} & \chl \textbf{64.7} & \chl \textbf{75.4} & \chl \textbf{79.5} & \chl \textbf{95.3} & \chl \textbf{89.8} & \chl \underline{94.7} & \chl \textbf{89.3} \\
\bottomrule
\end{tabular}
}
\label{compare_SOTA_table}
\end{table*}

\subsection{Comparison with State-of-the-Art Methods}

To validate the effectiveness of our proposed method, we compare with state-of-the-art unsupervised, semi- and full-supervised VI-ReID methods. The comparison results are summarized in Table \ref{compare_SOTA_table} and Table \ref{compare_llcm}. 

\noindent \textbf{\textit{Comparison with unsupervised VI-reID methods.}} As shown in Table \ref{compare_SOTA_table}, our method consistently outperform other state-of-the-art methods on both SYSU-MM01 and RegDB dataset. Compared to the recent well-performing method ASM~\cite{CAM2025}, our method improves Rank-1 accuracy by $0.9\%$ on SYSU-MM01. On RegDB, our method also demonstrates clear performance gain over previous best-performing method RPNR~\cite{RPNR}, improving Rank-1 and mAP by $4.4\%$ and $4.9\%$ respectively for Visible-to-Thermal search.

\begin{table}[hb]
	\begin{center}
		\caption{Comparison with state-of-the-art methods on LLCM dataset. The best unsupervised performance is marked in \textbf{Bold} and the second best is in \underline{underline}.}
		\resizebox{\linewidth}{!}{
			\begin{tabular}{c|c|c|cc|cc}
				\bottomrule
				\multicolumn{3}{c|}{Settings}&\multicolumn{2}{c|}{VIS to IR}&\multicolumn{2}{c}{IR to VIS}\\
				\hline
				Type& Method & Venue& Rank-1& mAP& Rank-1& mAP\\
				\hline
				\multirow{5}{*}{Supervised}
				& AGW~\cite{AGW} & TPAMI'21 & 51.5 & 55.3 & 43.6 & 51.8 \\
				& LbA~\cite{LbA} & ICCV'21 & 50.8 & 55.6 & 43.8 & 53.1 \\
				& CAJ~\cite{CAJ} & ICCV'21 & 56.5 & 59.8 & 48.8 & 56.6 \\
				& DEEN~\cite{DEEN} & CVPR'23 & 62.5 & 65.8 & 54.9 & 62.9 \\
				& CSDN~\cite{yu2025csdn} & TMM'25 & 63.7 & 66.5 & 55.8 & 63.5 \\
				\midrule
				\multirow{2}{*}{Semi-sup}
				& OTLA~\cite{OTLA} & ECCV'22 & 44.2 & 48.2 & 36.2 & 42.2 \\
				& HECC~\cite{zhang2025weakly} & ICCV'25 & 55.3 & 58.7& 47.3 & 53.3 \\
				\midrule
				\multirow{3}{*}{Unsupervised} 
				&ADCA~\cite{ADCA} &MM'22 &42.5 &46.9 &38.4 &44.4\\
			    &PGM~\cite{PGMAL}&CVPR'23 & \underline{44.9} &  \underline{49.0} & \underline{39.4} & \underline{45.3} \\
				& \textbf{Ours} & This paper & \textbf{52.2} & \textbf{56.1} & \textbf{45.3} & \textbf{51.2} \\
				\bottomrule
			\end{tabular}
		}
		\label{compare_llcm}
	\end{center}
\end{table}

Meanwhile, we also provide the performance of our method under the framework of a recent method BMIL~\cite{wang2026modality} which designs modality-agnostic clustering to estimate global instance-level relations. As observed in Table \ref{compare_SOTA_table}, the combination with BMIL (Ours$^\star$) leads to further improved accuracy, indicating the compatibility of our method with different frameworks.

\begin{table*}[ht]
	\centering
	\caption{Ablation study on SYSU-MM01 and RegDB.}
	\scalebox{0.98}{
	\begin{tabular}{c|cccc|cc|cc|cc}
	\bottomrule
                ~ &\multicolumn{4}{c|}{Components} & \multicolumn{2}{c|}{SYSU-MM01 (All Search)} & \multicolumn{2}{c|}{SYSU-MM01 (Indoor Search)} & \multicolumn{2}{c}{RegDB (Vis2Thermal)}\\
                \hline
               ~  &  $\mathcal{L}_{intra}$ & $\mathcal{L}_{cross}$  & $\mathcal{L}_{unified}$   &  $\mathcal{L}_{sd}$  & R1 & mAP & R1 & mAP & R1 & mAP \\
               \hline
               $\mathcal{M}1$ & \checkmark  & ~  & ~ & ~                                  & 39.4  & 38.7  & 47.6  & 56.4  & 47.6 & 45.1 \\
               $\mathcal{M}2$  & \checkmark & \checkmark & ~  & ~                 & 59.0 & 56.1 & 66.4 & 70.9  & 84.8 & 79.6 \\
                \hline 
               $\mathcal{M}3$  & \checkmark  & ~  & \checkmark & ~                & 60.1 & 57.6 & 67.5 & 72.6  & 91.7 & 86.1 \\
               $\mathcal{M}4$  & \checkmark  & \checkmark & ~ & \checkmark     & 59.4 & 57.2 & 65.6 & 71.4  & 90.7 & 84.9 \\
               $\mathcal{M}5$  & \checkmark  & ~  & ~ &  \checkmark               & \underline{65.4} & \textbf{62.3} & \underline{72.4} & \underline{76.8}  & \underline{94.7} & \underline{89.3} \\
               \rowcolor{mygray}  
               $\mathcal{M}6$  & \checkmark & ~ & \checkmark & \checkmark       & \textbf{66.0} & \underline{62.6} & \textbf{72.7} & \textbf{76.8}  & \textbf{95.3} & \textbf{89.6} \\
         \bottomrule              
	\end{tabular}
        }
\label{ablation_table}
\end{table*}

On LLCM dataset, our method also shows superior performance compared to other unsupervised VI-reID methods. Specifically, our method surpasses PGM by $7.3\%$ and $5.9\%$ on the Rank-1 Accuracy of Visible-to-Infrared and Infrared-to-Visible evaluation. The comparisons prove the effectiveness of our proposed method, despite its simplicity in both concept and implementation.

\noindent \textbf{\textit{Comparison with semi- and fully-supervised VI-reID methods.}} Compared to semi-supervised VI-reID with intra-modality annotations, our method is also competitive, surpassing both OTLA and DPIS, while further reducing the performance margin with the recent SoTA semi-supervised method HECC. Since our method mainly focus on cross-modality learning, the proposed components can potentially be applied to improve the cross-modality learning in semi-supervised VI-reID. Additional comparison with fully-supervised methods indicate that our method further narrows the performance gap with supervised counterparts, demonstrating the potential of unsupervised VI-reID under proper learning design.

\subsection{Ablation Study}
In Table \ref{ablation_table}, we present the ablation results on SYSU-MM01 and RegDB dataset, to investigate the impact of each proposed component: Modality-unified Prototypical Contrast and Prototype-guided Self-distillation.

\noindent \textbf{\textit{Advantage of Modality-unified Prototypical Contrast.}}
First, by comparing $\mathcal{M}2$ and $\mathcal{M}3$ in Table \ref{ablation_table}, we observe that the modality-unified contrastive loss outperforms the cross-modality loss~\cite{PGMAL} on both SYSU-MM01 and RegDB. Considering that $\mathcal{M}2$ and $\mathcal{M}3$ rely on the same association strategy, \textit{i.e.} optimal transport based cluster matching, it proves the modality-unified optimization is indeed able to facilitate learning more discriminative cross-modality representation. Further, by comparing $\mathcal{M}4$ and $\mathcal{M}6$, we see that the cross-modality loss is not suitable to be combined with online unified distillation, probably due to optimization conflict when enforcing local cross-modality contrast without referring to intra-modality similarity. 

\noindent \textbf{\textit{Effectiveness of Prototype-guided Self-distillation.}} 
Compared to $\mathcal{M}1$-$\mathcal{M}3$, the performance with only prototype-guided online self-distillation ($\mathcal{M}5$) demonstrates a clear advantage; For instance, the Rank-1 (All Search) accuracy improves over $\mathcal{M}2$ by $6.4\%$, and the margin is enlarged to $9.9\%$ on RegDB. This indicates our prototype-guided online self-distillation is able to work well when utilized alone, demonstrating its robustness for facilitating cross-modality instance-prototype similarity optimization, without relying on the offline cluster matching.

\noindent \textbf{\textit{Effectiveness of the combination of proposed components.}}
Finally, we integrate both Modality-unified Prototypical Contrast Loss and Prototype-guided Self-distillation Loss, which leads to $\mathcal{M}6$. It can be seen that even when $\mathcal{M}5$ already produces impressive result, adding offline association based Modality-unified Prototypical Contrast further boosts the performance, increasing Rank-1 by $0.6\%$ on SYSU-MM01 and $0.6\%$ on RegDB. The performance ablation proves the complementarity of offline association and our online distillation when optimized under the same modality-unified framework.

\subsection{Parameter Analysis}
In this subsection, we analyze the impact of teacher temperature $\gamma$ for Prototype-guided Self-distillation. The model performance with different values of $\gamma$ is illustrated in Figure \ref{fig_analyze}(a). From the figure, we can see that the model achieves the best performance when $\gamma$ is set to $0.05$. When increasing the distill temperature, the resulting teacher distribution becomes softer, which may result in the negative prototypes being assigned a large distill similarity, potentially harming model learning. On the other hand, when a smaller distill temperature is utilized, the resulting teacher distribution may get too sharp, causing ambiguous positive prototypes to be suppressed. Therefore, we set the teacher temperature $\gamma$ to 0.05 in our experiments.

\subsection{Further Analysis on Prototype-guided Self-distillation}

To further look into the proposed Prototype-guided Self-distillation, we provide additional experiments that replace the proposed design with other possible alternatives for comparison.

\begin{table}[hb]
	\centering
	\caption{Further analysis into the prototype-guided self-distillation, on SYSU-MM01 dataset.}
	\scalebox{0.88}{
	\begin{tabular}{cc|cc|cc}
	\bottomrule
                \multicolumn{2}{c|}{Components} & \multicolumn{2}{c|}{All search} & \multicolumn{2}{c}{Indoor search} \\
                \hline
              Unified Distill.  &  Instance-adapt. Fusion & R1 & mAP & R1 & mAP \\
               \midrule
               \xmark & \cmark                & 63.4 & 61.4 & 72.1 & 76.9 \\
                \cmark  & \xmark               & 65.0 & 61.8 & 72.6 & 76.8  \\
                \cmark & \cmark       & 66.0 & 62.6 & 72.7 & 76.8  \\
        \bottomrule                
	\end{tabular}
        }
\label{analyze_table}
\end{table}

\noindent \textit{\textbf{Impact of modality-unified distillation.}} Specifically, we first present the results of replacing the modality-unified distillation with the combination of intra-modality and cross-modality self-distillation. As shown in Table \ref{analyze_table}, when the separate distillation is adopted instead of unified distillation, the performance of All Search suffers from a significant decrease. While the Indoor Search accuracy remains stable, the overall decreased performance suggests that modality-unified optimization is crucial to the effectiveness of Prototype-guided Self-distillation. Enforcing separate intra-modality and cross-modality distillation cannot jointly attend to the similarity discrepancy across modality, thus weakening the effect of distillation.

\noindent \textit{\textbf{Necessity of instance-adaptive similarity fusion.}} In addition, we also investigate how much the \textit{instance-adaptive similarity fusion} contributes to the distillation. In Table \ref{analyze_table}, it can be observed that when replacing the instance-adaptive similarity fusion with simply the prototype-to-prototype similarity as the distillation source, the performance exhibits a noticeable drop, proving that our proposed instance-adaptive similarity fusion is indeed necessary for coping with the intra-cluster noise. Without considering the instance-to-prototype similarity, the distillation tends to ignore the semantic inconsistency between instances and its centroid prototype, thereby compromising the distillation effect.

 \begin{figure}[ht]
\centering
\begin{subfigure}{0.227\textwidth}
\centering
\includegraphics[width=1.0\textwidth]{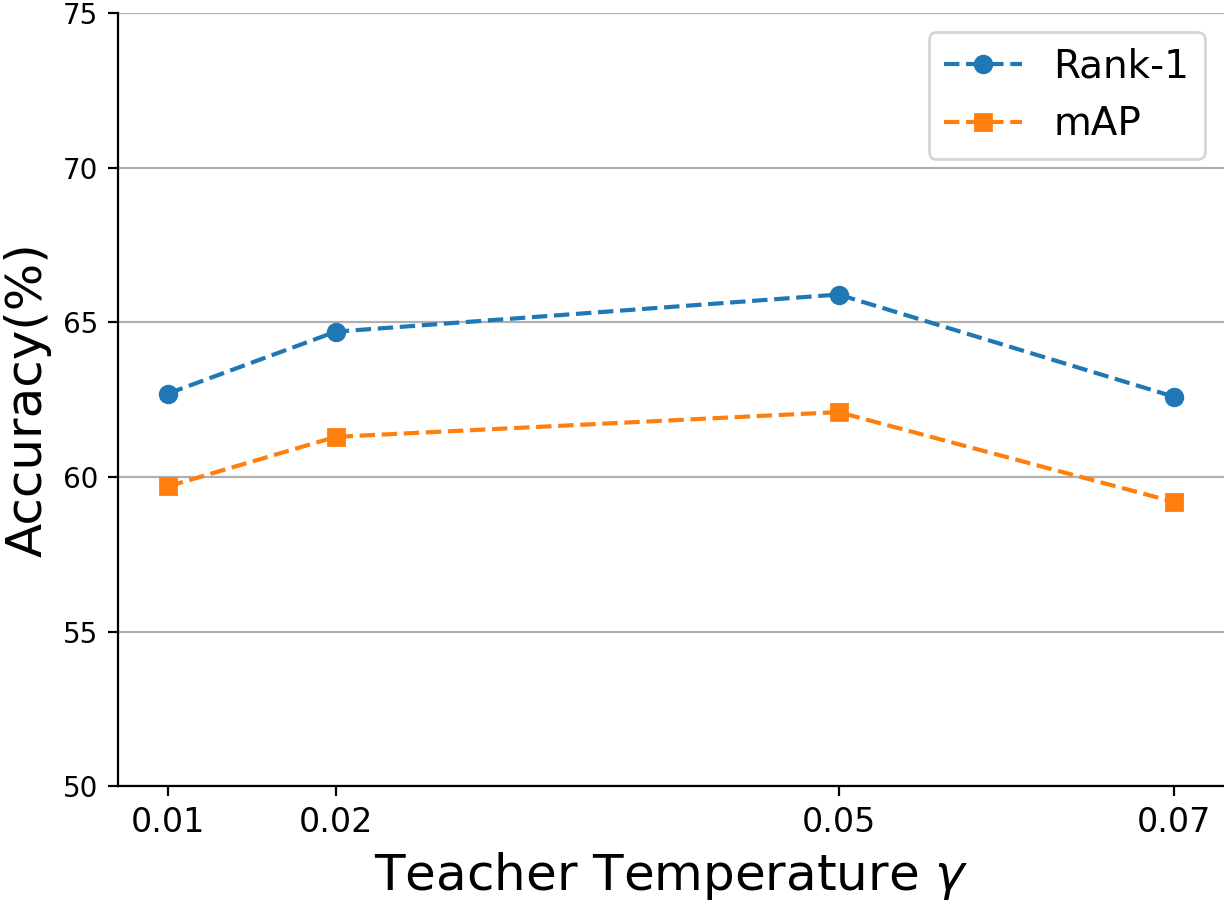} 
\caption{Analysis on $\gamma$}
\end{subfigure}
\quad 
\begin{subfigure}{0.225\textwidth}
\centering
\includegraphics[width=1.0\textwidth]{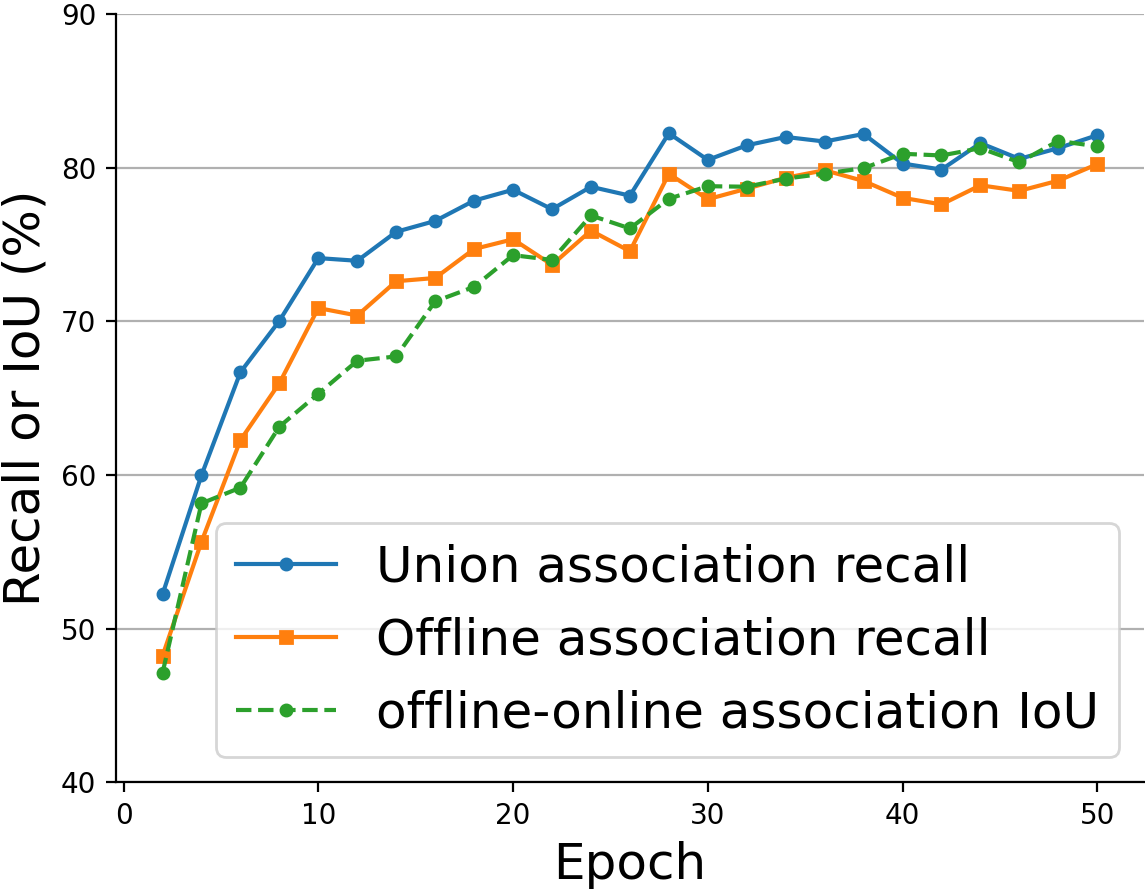} 
\caption{Analysis of association}
\end{subfigure}
\caption{Analysis on teacher temperature $\gamma$, and comparison of offline and online distillation.}
\label{fig_analyze}
\end{figure}

\noindent \textit{\textbf{Complementarity of offline association and online distillation.}}
To better understand the dynamics of \textit{optimal transport based offline association} and \textit{prototype-guided online distillation}, we analyze the association statistics during training, including: 

\noindent 1) \textit{Recall of offline association:} the ratio of "\textit{positive prototypes}" that are correctly retrieved by offline association. Specifically, "\textit{positive prototypes}" are computed by assigning the label of the most dominant class in each cluster as the prototype "\textit{ground truth}" label;  

\noindent 2) \textit{Recall of the union of offline and online distillation.} Since online distillation assigns soft label, we take the \textit{Top-2} prototypes with the highest soft similarity as the online associated prototypes, then compute the recall of offline-online association union.

\noindent 3) \textit{Intersection-over-Union (IoU) between offline association and online distillation}: reflecting the coincidence degree of them.
 
 From Figure \ref{fig_analyze}(b), it can be observed that: First, the recall of both offline association and union association are continuously increasing as training proceeds, suggesting the association quality gets gradually improved during training. After around 30 epochs, the association recall becomes stable. 
 
Second, the recall of union association consistently outperforms the recall of offline association, indicating that online distillation is able to retrieve missing positive prototypes that enhances representation learning. Moreover, the IoU between offline association and online distillation gradually ramps up during training. It proves their coincidence degree is low at beginning which offers complementarity supervision signals, and the two types of association become more consistent as model converges.

\subsection{Visualization}

\begin{figure}[ht]
\centering
\begin{subfigure}{0.225\textwidth}
\centering
\includegraphics[width=1.0\textwidth]{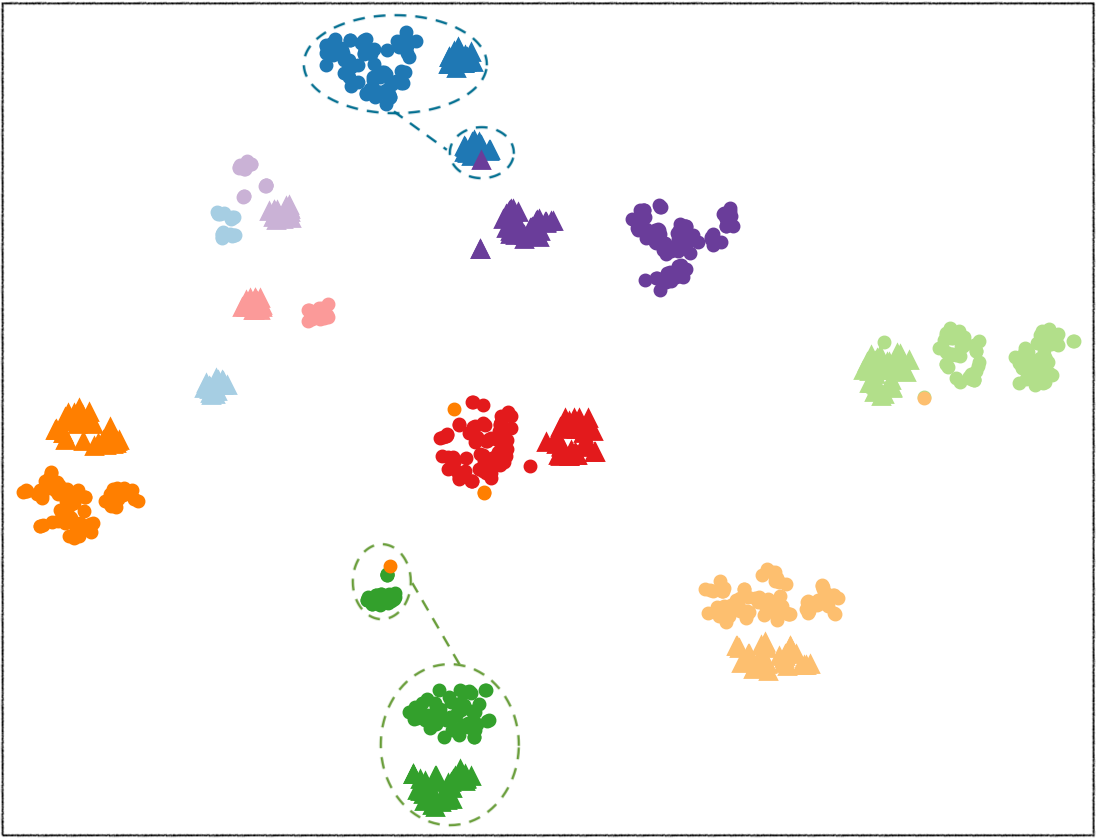} 
\caption{Baseline}
\end{subfigure}
\quad 
\begin{subfigure}{0.225\textwidth}
\centering
\includegraphics[width=1.0\textwidth]{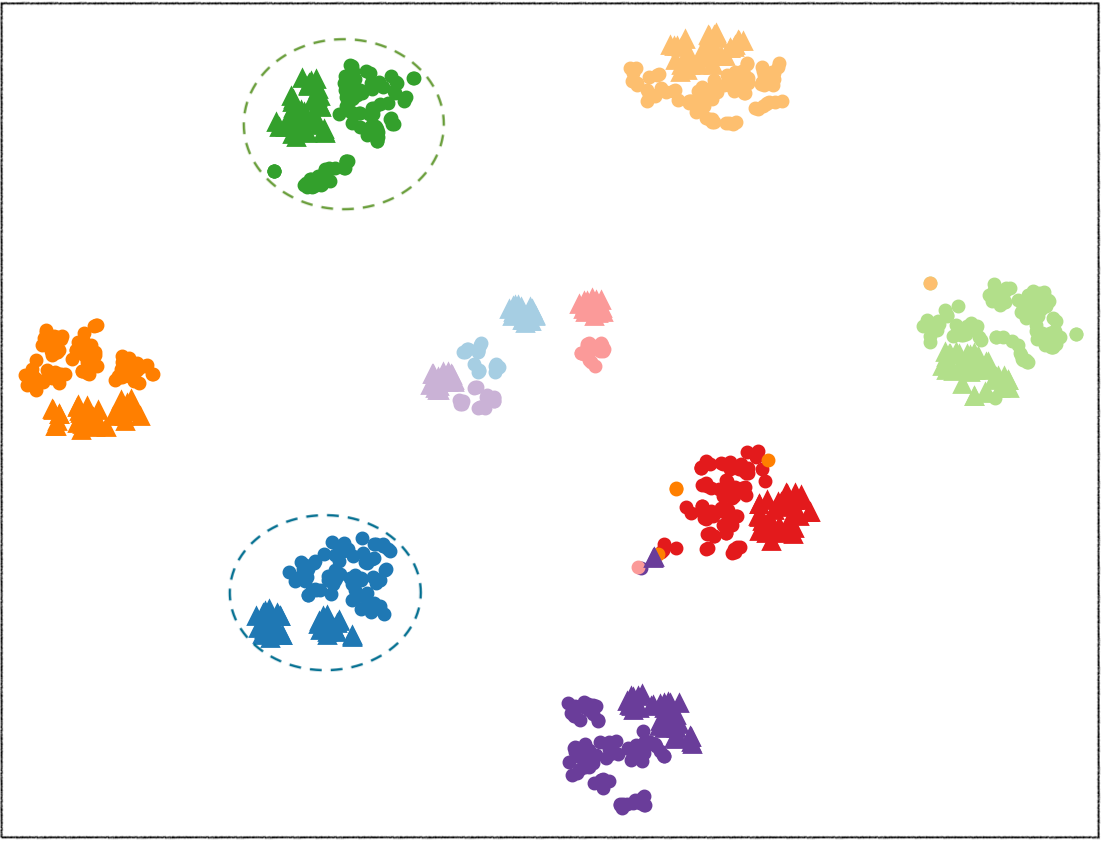} 
\caption{Our method}
\end{subfigure}
\caption{T-SNE feature visualization of images from 10 randomly selected identities of SYSU-MM01 training set. Different color indicates different identity. Circle and triangle denote visible and infrared image, respectively.}
\label{fig_visualize}
\end{figure}

To intuitively analyze the feature improvement, we present the visualization of features using the baseline and our method in Figure \ref{fig_visualize}. From the figure, we observe that our model generates more compact intra-class distributions compared to the baseline. For instance, features of the identities marked in blue or green are dispersed into a number of sub-clusters by the baseline model, while our method effectively associates the intra-class features together and forms tighter distribution. The comparison indicates that our method produces more discriminative features that facilitates identity recognition.

\section{Conclusion}
In this paper, we have proposed a simple yet effective method for unsupervised visible-infrared person re-ID. By identifying the inefficiency of optimal transport based cluster matching and cross-modality loss, we propose a unified learning framework to jointly optimize the contrastive similarity of all modalities, thus explicitly addressing the modality discrepancy and achieve better modality invariance. By uniquely re-purposing the centroid prototype as a steady yet informative online teacher, we further design prototype-guided self-distillation to rectify the online instance-prototype similarity relation, so as to complement offline association and provide stronger supervision for modality-unified contrastive learning. The effectiveness of our method and the proposed components are validated by our extensive experiments and analysis on benchmark VI-reID datasets.

\begin{acks}
This paper is supported by the General Project of Basic Science Research in Higher Education Institutions of Jiangsu Province, China (No. 25KJD520008).
\end{acks}

\bibliographystyle{ACM-Reference-Format}
\balance
\bibliography{reference}

\end{document}